\documentclass{article}

\PassOptionsToPackage{numbers}{natbib}
\usepackage[preprint]{neurips_2024}

\usepackage[utf8]{inputenc} 
\usepackage[T1]{fontenc}    
\usepackage{hyperref}       
\usepackage{url}            
\usepackage{booktabs}       
\usepackage{amsfonts}       
\usepackage{nicefrac}       
\usepackage{microtype}      
\usepackage{xcolor}         

\title{How Much Data Is Enough? \\Uniform Convergence Bounds for \\Generative \& Vision-Language Models \\under Low-Dimensional Structure}

\author{%
  Paul M. Thompson\thanks{Author can be reached by email at pthomp at usc dot edu} \\
  Stevens Institute for Neuroimaging and Informatics, University of Southern California\\
  Los Angeles, CA 90033 \\
  \texttt{pthomp@usc.edu} \\
}

\usepackage{graphicx}
\begin{document}

\maketitle

\begin{abstract}
  Modern generative and vision–language models (VLMs) are increasingly used to support decisions in scientific and medical settings, where predictive probabilities must be not only accurate but also well calibrated. While such models often achieve strong empirical performance with moderate data, it remains unclear when their predictions can be expected to generalize uniformly across inputs, classes, or subpopulations, rather than only on average. This distinction is particularly important in biomedical applications, where rare conditions or specific subgroups may exhibit systematically larger errors despite low overall loss.

We study this question from a finite-sample perspective and ask: under what structural assumptions can generative and vision–language models achieve uniformly accurate and calibrated behavior with practical sample sizes? Rather than analyzing general model parameterizations, we focus on induced families of classifiers obtained by varying prompts or semantic embeddings within a restricted representation space. We show that when model outputs depend smoothly on a low-dimensional semantic representation—an assumption empirically supported by the spectral structure of text and joint image–text embeddings—classical uniform convergence arguments yield meaningful, non-asymptotic guarantees.

Our main results establish finite-sample uniform convergence bounds for accuracy and calibration functionals of VLM-induced classifiers under Lipschitz stability with respect to prompt embeddings. The resulting sample complexity depends on the intrinsic or effective dimension of the embedding space, rather than its ambient dimensionality. We further derive spectrum-dependent bounds that make explicit how eigenvalue decay in the embedding covariance governs data requirements. These results help explain why VLMs can generalize reliably with far fewer samples than would be predicted by parameter count alone.

Finally, we discuss practical implications for training and evaluating generative and vision–language models in data-limited biomedical settings. In particular, our analysis clarifies when current dataset sizes are sufficient to support uniformly reliable predictions, and why commonly used average calibration metrics may fail to detect worst-case miscalibration. Together, these results provide a principled framework to reason about data sufficiency, calibration, and reliability in modern generative and multimodal models.
\end{abstract}

\section{Why does this matter?}

Many people are training generative models to create high-quality images and videos, and it is important to know how much data is needed to create an accurate generative model. In medical applications of generative models and VLMs (Boecking 2022, Wang 2022), the generated outputs need to be accurate rather than simply visually pleasing. Some formal criteria include accuracy, and also calibration (Guo 2017; Vaicenavicius 2019; Kumar 2019, Dhinagar 2025a,b,c). In a generative setting, accuracy reflects whether classifiers derived from the model, or decisions based on conditional distributions, correctly predict observed labels. Calibration ensures that the probabilities produced by the model are statistically meaningful, so that events predicted with probability $p$ occur approximately a fraction $p$ of the time. Some recent works have focused on VLMs that align text and images, with surprising classification accuracy even at moderate amounts of data. Some of these models perform surprisingly well even versus custom-trained image classifiers.   

Empirically, modern generative and vision-language models often exhibit strong classification performance even in low- or moderate-data regimes (Radford 2021). However, these successes raise a deeper theoretical issue: while pointwise accuracy may be high on benchmark tasks, it remains unclear when such models can be expected to generalize uniformly across inputs and classes, or to produce well-calibrated probabilities rather than overconfident predictions (Grisel 2025). Understanding when and why these properties emerge requires us to move beyond asymptotic arguments to finite-sample guarantees.

This raises a fundamental question: under what structural assumptions can modern generative and vision-language models achieve uniformly accurate and calibrated behavior with finite, practically attainable sample sizes?

To reason about accuracy and calibration simultaneously, one needs probabilistic bounds that control deviations between empirical and true distributions uniformly over relevant function classes. Here, “uniformly” means that the model’s error is controlled in the worst case across subjects, brain regions, or clinical categories, rather than only “on average” across the dataset. In other words, uniformly refers to controlling the maximum deviation between predicted and true outcomes across all inputs of interest, so that no subgroup, image type, or disease category exhibits systematically larger errors. 

This perspective aligns with recent remarks on domain-wide error by Varoquaux (2025), where he emphasized the need to assess model error and uncertainty across the entire input domain rather than rely solely on average performance metrics. Such domain-wide evaluation is important in medical and neuroscience applications, where models with low average error can still exhibit clinically meaningful failures on specific subpopulations or rare conditions. Uniform accuracy and calibration aim to provide robustness guarantees that guard against these kinds of failures.

Classical uniform convergence results provide a principled way to formalize this notion of domain-wide reliability by bounding the maximum discrepancy between empirical and true behavior across all relevant inputs. These ideas form a natural starting point to understand when (and how much) generative and vision-language models can be trusted beyond visual fidelity alone.

Sections 2, which follows, provides a classical motivation and terminology. Our main technical contributions are in Sections 3 to 5, where we derive finite-sample uniform convergence bounds for VLM-induced classifiers under a low-dimensional prompt structure. Section 6 discusses practical consequences of our work and future directions. 

\section{Related Work}

\subsection{The Dvoretzky--Kiefer--Wolfowitz (DKW) Inequality (Classical Uniform Convergence) --- why the worst-case error shrinks like \texorpdfstring{$1/\sqrt{n}$}{1/sqrt(n)}}

Suppose $X_1, X_2, \ldots, X_n$ are independent and identically distributed (i.i.d.) random samples from some unknown distribution $F$.

The true cumulative distribution function (CDF) $F(x)$ gives the probability that a random draw is less than or equal to $x$.

The empirical cumulative distribution function (empirical CDF) $F_n(x)$ is computed from the data and is defined as the fraction of samples that are less than or equal to $x$:
\[
F_n(x) \;=\; \frac{1}{n} \sum_{i=1}^n \mathbf{1}\{X_i \le x\},
\]
where $\mathbf{1}\{X_i \le x\}$ is an indicator function that equals $1$ if $X_i \le x$ and $0$ otherwise.

Now the result:

For every $\varepsilon > 0$, the probability that the empirical cumulative distribution function deviates from the true cumulative distribution function by more than $\varepsilon$ at any point on the real line is bounded by
\[
\Pr\!\left( \sup_{x \in \mathbb{R}} \left| F_n(x) - F(x) \right| > \varepsilon \right)
\;\le\;
2\,\exp\!\left(-\,2\,n\,\varepsilon^2\right).
\]

That is, the probability that the largest difference between the empirical and true distributions exceeds $\varepsilon$ decays exponentially fast with the sample size $n$.

 This inequality was first proven in 1956 by Aryeh Dvoretzky, Jack Kiefer, and Jacob Wolfowitz; in 1990, Pascal Massart (Massart 1990) showed the “sharp” version with the constant 2 on the right-hand side.
 
At a high level, the Dvoretzky-Kiefer-Wolfowitz (DKW) result answers a simple question: \emph{how many samples do we need before the distribution we observe in data reliably matches the true, underlying distribution?} Rather than focus on average error, it guarantees that the largest possible discrepancy between what we see in data and what actually exists in the population is small, with high probability. 

What makes this result remarkable is that it was solved in the 1950s by Aryeh Dvoretzky, Jack Kiefer, and Jacob Wolfowitz, before modern machine learning existed. They showed that, regardless of the underlying distribution, the maximum error between the empirical and true distributions shrinks at a universal rate as more data are collected. In other words, the bound does not depend on the complexity of the distribution itself-only on how much data we have. In other words, if we collect more data, we can be increasingly confident that the empirical distribution is close to the true distribution everywhere, not just on average. 

We can equivalently express the Dvoretzky--Kiefer--Wolfowitz inequality in $(\varepsilon,\delta)$ form by choosing a desired confidence level $\delta \in (0,1)$ and solving for $\varepsilon$. The result is that, with probability at least $1-\delta$,
\begin{equation}
\sup_{x \in \mathbb{R}} \bigl| F_n(x) - F(x) \bigr|
\;\le\;
\sqrt{\frac{1}{2n}\,\ln\!\left(\frac{2}{\delta}\right)}.
\end{equation}

Equivalently, the DKW inequality states that with confidence $1-\delta$, the empirical distribution uniformly approximates the true distribution to within an error $\epsilon(n,\delta)$ that decreases as the inverse square root of the sample size. DKW controls the maximum mismatch anywhere on the probability axis, and it applies to one scalar random variable. An alternative measure of calibration, the Expected Calibration Error (ECE) can be viewed as a coarse, binned approximation to the same distributional mismatch that DKW controls uniformly. While ECE summarizes calibration error on average across probability bins, DKW provides a guarantee that no bin-or any finer subdivision of the probability range-exhibits large error with high confidence. As a result, a model may have low ECE yet still be dangerously miscalibrated for rare conditions or extreme confidence levels. As such, ECE measures average calibration but DKW explains when calibration is reliable everywhere. Low ECE does not imply small DKW deviation, but small DKW deviation does imply small ECE (up to bin width).

DKW is powerful, but only controls how well the empirical distribution of one random variable approximates its true distribution, uniformly. A classifier can be viewed as a rule that induces a family of one-dimensional distributions-one for each possible input, subgroup, or confidence level. Learning a classifier or a calibrated predictor is equivalent to learning a conditional distribution $P(Y = 1 \mid X = x)$, whose behavior must be reliable across the input domain. In practice, a classifier or generative model induces a collection of distributions, indexed by inputs, subgroups, or probability thresholds. Uniform reliability therefore requires controlling errors not for one distribution, but simultaneously across a family of related distributions. Glivenko–Cantelli theory provides the mathematical framework for this setting, characterizing when empirical estimates converge uniformly to their true values across entire classes of functions or conditionals, rather than for a single fixed distribution.

In what follows, Glivenko–Cantelli theory is used as a conceptual benchmark rather than an assumption. That is, we treat uniform convergence as a desirable property and examine whether modern generative and vision–language models, under structural constraints such as low effective dimension or restricted embeddings, can plausibly satisfy Glivenko–Cantelli–type guarantees for accuracy and calibration. Rather than asserting that modern generative models satisfy Glivenko–Cantelli conditions, we use the theory as a way to ask under what structural assumptions such uniform guarantees might hold.

\subsection{Glivenko–Cantelli Property ($\varepsilon,\delta$ Definition)}

Let $(X_1,Y_1), \ldots, (X_n,Y_n)$ be independent samples drawn from an unknown distribution $\mathcal{D}$, and let $\mathcal{F}$ be a class of real-valued functions defined on $(X,Y)$. We say that $\mathcal{F}$ has the \emph{Glivenko--Cantelli property} if, for every $\varepsilon > 0$ and $\delta \in (0,1)$, there exists a sample size $n(\varepsilon,\delta)$ such that, for all $n \ge n(\varepsilon,\delta)$,
\begin{equation}
\Pr\!\left(
\sup_{f \in \mathcal{F}}
\left|
\frac{1}{n}\sum_{i=1}^n f(X_i,Y_i)
-
\mathbb{E}_{(X,Y)\sim\mathcal{D}}[f(X,Y)]
\right|
>
\varepsilon
\right)
\;\le\;
\delta.
\end{equation}

If the Glivenko--Cantelli property holds for a model, then with high confidence $1-\delta$, empirical estimates computed from finite data approximate their true population values to within $\varepsilon$, uniformly across all functions in the class. In this definition, the function class $\mathcal{F}$ represents performance or calibration functionals of model predictions (e.g., loss or calibration error), rather than the underlying data distribution itself. It is therefore of interest to ask whether this property holds for a given model class (e.g., all linear classifiers in a latent space, VAE decoders with fixed architecture, all vision--language model classifiers induced by prompt templates, or all conditional distributions $p_\theta(Y \mid X)$ realizable by a model family), and for which classes of evaluation functions (e.g., loss functions such as $0$--$1$ loss or cross-entropy, reconstruction loss, calibration bin indicators used in expected calibration error, subgroup-specific error functions, calibration error of conditional predictions, or tail-risk and worst-case error functionals).

In this framework, the \emph{hypothesis class} refers to the family of predictive or generative models under consideration, such as classifiers induced by a vision--language model or conditional distributions defined by a variational autoencoder. The \emph{evaluation class}, by contrast, consists of performance and calibration functionals computed from model predictions, including loss, accuracy indicators, and calibration discrepancies. Glivenko--Cantelli theory applies to the evaluation class, ensuring that empirical estimates of these quantities are uniformly reliable across all models and evaluation functions under consideration.

\subsection{Relation to the Hungarian Lemma (Strong Approximation)}

A closely related line of work is the so-called \emph{Hungarian construction} or \emph{Hungarian coupling}, developed in a series of papers between 1975 and 1976 by a group of Hungarian probabilists, notably J\'anos Koml\'os, P\'al Major, and G\'abor Tusn\'ady (and therefore sometimes referred to as the KMT ``strong approximation''; Koml\'os et al., 1975, 1976). Their result shows that the empirical distribution function can be coupled almost perfectly with a Brownian bridge, yielding a pathwise approximation that is sharper than the Dvoretzky--Kiefer--Wolfowitz (DKW) inequality and holds uniformly over the entire domain.

While DKW and Glivenko--Cantelli theory provide probabilistic guarantees on worst-case deviations, the Hungarian construction strengthens these results by showing that empirical and true distributions can be matched point-by-point up to logarithmic factors. Conceptually, this work reinforces the same central idea: that empirical distributions stabilize uniformly with sufficient data, forming a key bridge between classical probability and modern uniform convergence theory.

\paragraph{Hungarian Construction.}
Let $X_1, \ldots, X_n$ be independent samples drawn from a continuous distribution with cumulative distribution function $F$, and let $F_n$ denote the empirical cumulative distribution function computed from the data. Define the empirical fluctuation
\begin{equation}
\sqrt{n}\,\bigl(F_n(x) - F(x)\bigr),
\end{equation}
which measures how far the empirical distribution deviates from the truth at each point $x$.

The Hungarian construction shows that this empirical fluctuation can be matched, point by point, to a simple continuous stochastic process known as a \emph{Brownian bridge}. A Brownian bridge is a random, continuous curve that starts at zero, ends at zero, and fluctuates randomly in between. It behaves like random noise, but is constrained to return to zero at the endpoints, making it a natural model for centered cumulative fluctuations.

Formally, the Hungarian construction states that one can place the empirical distribution and a Brownian bridge on the same probability space such that, with high probability,
\begin{equation}
\sup_{x \in \mathbb{R}}
\left|
\sqrt{n}\,\bigl(F_n(x) - F(x)\bigr) - B\!\left(F(x)\right)
\right|
\;\le\;
C \log n,
\end{equation}
where $B(\cdot)$ denotes a Brownian bridge and $C$ is a constant.

The DKW inequality, the Glivenko--Cantelli property, and the Hungarian lemma all address the same fundamental question: \emph{how close is the empirical world to the true one, everywhere all at once?} We note that flow-based models used in generative AI are conceptually similar to the Hungarian construction in that they attempt to transform complex distributions into simple reference distributions, but they do not enjoy uniform convergence bounds of this type. Indeed, the Hungarian lemma is inherently one-dimensional, and there is no direct mathematical extension of it to neural flows.

The Hungarian construction achieves something modern AI models would ideally possess: uniform, sup-norm control over the entire domain, rather than convergence in expectation or likelihood alone. In this work, however, we do not assert such guarantees in general. Instead, we seek to derive uniform convergence bounds for vision--language models under explicit structural assumptions.

\subsection{Sample Complexity}

We conclude this section by defining \emph{sample complexity}, which will play a central role in the analysis that follows. Sample complexity refers to the number of independent samples required to guarantee, with confidence $1-\delta$, that empirical estimates of performance or calibration deviate from their true population values by at most $\varepsilon$, uniformly across a specified class of models or evaluation functions. We use $n(\varepsilon,\delta)$ to denote the number of samples required to achieve uniform accuracy and calibration guarantees at tolerance $\varepsilon$ and confidence $1-\delta$.

This quantity appears implicitly in both the DKW inequality and Glivenko--Cantelli theory, where sample complexity corresponds to the minimal $n(\varepsilon,\delta)$ for which uniform convergence holds.

\subsection{Sample Complexity in Generative Models}

Sample complexity plays a central role in determining whether uniform accuracy and calibration guarantees are attainable in practice. While classical results such as DKW and Glivenko--Cantelli theory characterize how sample size controls uniform deviations for simple distributional objects or function classes, extending these insights to modern generative models is challenging.

\subsubsection{Related Applications: Low-Dimensional Structure in Face Recognition}

Empirical evidence from computer vision suggests that meaningful data often concentrate near low-dimensional manifolds embedded in high-dimensional ambient spaces. A prominent example is face recognition, where variation is frequently attributed to a small number of underlying factors (e.g., identity, pose, illumination, and expression; Turk and Pentland, 1991; Belhumeur et al., 1997). A long line of work has shown that models exploiting this low-dimensional structure, either explicitly through subspace methods or implicitly through learned embeddings, can generalize well with relatively modest amounts of data.

These observations motivate the hypothesis that intrinsic dimension, rather than ambient dimension, governs sample complexity in practice. For classical models such as low-rank principal component analysis, explicit finite-sample bounds are available that scale with rank and noise level rather than input dimensionality, yielding uniform convergence guarantees for reconstruction error and related quantities.

At present, however, there are no general results characterizing the sample complexity of modern generative models---such as variational autoencoders or vision--language models---in terms of their intrinsic or effective dimension that yield uniform, finite-sample guarantees for accuracy and calibration.

Prior theoretical work has shown that for autoencoders and variational autoencoders, finite-sample generalization bounds can be derived under explicit assumptions on the Jacobian of the encoder--decoder map. In particular, when the Jacobian has low rank or rapidly decaying singular values, reconstruction error scales with the effective latent dimension rather than the ambient input dimension. These bounds arise from locally linearizing the generative map and show that error depends on the spectrum of the Jacobian: large singular values amplify estimation error, while spectral decay yields favorable data-sufficiency laws. Existing results, however, focus primarily on reconstruction or likelihood-based objectives and do not provide uniform, finite-sample guarantees for downstream accuracy or probability calibration.

Related work has also developed generalization theory for neural networks trained by gradient descent by exploiting low-rank structure in the Jacobian of the input--output map. For many structured datasets, the Jacobian exhibits a small number of large singular values and many small ones, creating a low-rank ``information space'' and a high-dimensional ``nuisance space.'' Learning proceeds more efficiently in the information subspace, while directions associated with small singular values contribute disproportionately to overfitting unless controlled.

Although these results are not framed in terms of $(\varepsilon,\delta)$ sample complexity or uniform worst-case guarantees, they provide strong evidence that models may require far fewer samples than their parameter counts suggest when most signal resides in a low-dimensional subspace.

\section{Sample Complexity for Vision--Language Models}

Recent empirical work suggests that text embeddings produced by large language models, and by extension vision--language models, exhibit strong low-rank structure, with most variance concentrated in a small number of directions. This effect appears especially pronounced in medical and clinical settings, where disease concepts are highly structured, correlated, and hierarchically organized. As a result, the effective dimensionality of text embeddings used for disease description or prompting is often far smaller than their ambient dimension.

These observations motivate an intrinsic-dimension perspective on sample complexity for vision--language models (VLMs), in which generalization depends on the geometry of the learned representation space rather than the nominal dimensionality of the embeddings.

\subsection{Effective Dimension of Text Embeddings}

We assume that the joint data distribution over images and text embeddings is supported on, or concentrated near, a $d$-dimensional manifold embedded in $\mathbb{R}^D$, with $d \ll D$. Empirically, this manifests as rapid spectral decay in the covariance of text embeddings, indicating that only a small number of semantic directions capture most of the meaningful variation.

From a learning-theoretic perspective, this distinction is critical: uniform convergence rates and sample complexity depend on the intrinsic dimension $d$ governing variability in the representation, rather than the ambient embedding dimension $D$. This aligns with classical results for low-rank principal component analysis and manifold learning, and provides a plausible explanation for why VLMs can generalize effectively with moderate sample sizes despite operating in very high-dimensional spaces.

\subsection{Disease Classification: One-Hot Labels vs.\ Semantic Manifolds}

Traditional disease classification frameworks often represent diagnostic categories using one-hot encodings, treating diseases as unrelated and orthogonal labels. In such settings, sample complexity scales with the number of disease classes, and each class must be learned largely independently.

In contrast, diseases are not semantically independent. Many conditions share overlapping phenotypes, imaging signatures, or pathophysiological mechanisms, and thus naturally reside on a low-dimensional semantic manifold. Vision--language models exploit this structure by embedding disease descriptions into a shared latent space, where proximity reflects semantic relatedness.

As a consequence, disease classification using VLMs can be viewed as learning decision boundaries in a low-dimensional latent space, rather than fitting separate classifiers for each one-hot label. This shared structure substantially reduces effective complexity and, in turn, sample requirements.

\subsection{Sample Complexity in Low-Dimensional Latent Spaces}

When predictions are made in a latent space of dimension $d$, rather than $D$, classical arguments apply: the number of samples required to achieve a given accuracy or calibration tolerance scales with $d$, not with the ambient dimension. This applies both to linear classifiers defined in the latent space and to more general smooth decision functions whose complexity is controlled by the geometry of the embedding.

From this perspective, VLMs act as complexity-reducing maps, transforming high-dimensional sensory inputs into structured representations where learning is statistically efficient.

\subsection{Jacobians, Fisher Information, and Spectral Decay}

The effective dimension of a VLM can be characterized in several equivalent ways. One is through the Jacobian of the image--text mapping, whose rank and singular value spectrum quantify how many directions in representation space meaningfully influence predictions. Rapid spectral decay implies that perturbations outside a small subspace have negligible effect on outputs.

Closely related is the Fisher information, which captures the sensitivity of model predictions to changes in latent variables. When Fisher information is concentrated in a low-dimensional subspace, estimation error and generalization behavior are governed primarily by that subspace, again yielding sample complexity that scales with intrinsic rather than ambient dimension.

Empirically, both Jacobian spectra and Fisher information matrices derived from trained VLMs often exhibit strong low-rank structure, especially in domain-specific settings such as medical imaging, where semantic variability is constrained.

\subsection{Effective Dimension from Real Embeddings}

These theoretical considerations are supported by direct empirical measurements of effective dimension in real embedding spaces. Spectral analyses of text and joint image--text embeddings consistently show that a small number of principal components explain most of the variance, suggesting that VLMs operate in a regime where intrinsic dimension is modest even when embeddings are high-dimensional.

This empirical low-rank structure provides concrete justification for modeling assumptions that lead to favorable sample complexity bounds.

\subsection{Implications: A Shared, Low-Complexity Hypothesis Class}

Taken together, these observations suggest that vision--language models induce a shared, low-rank hypothesis class across diseases, rather than a collection of independent classifiers. By aligning images and text in a common semantic space, VLMs dramatically reduce effective complexity, enabling uniform accuracy and calibration to be achieved with fewer samples than would be required under one-hot or task-specific formulations.

In the remainder of this section, we formalize these intuitions and derive sample complexity bounds that make explicit how intrinsic dimension, spectral decay, and shared representations govern uniform convergence behavior in vision--language models.

\section{Arguments for Uniform Convergence for Vision--Language Models and Assumptions}

The uniform convergence arguments developed here rely on Lipschitz control of logits, probabilities, and calibration surrogates with respect to the prompt embedding. This condition is naturally satisfied by CLIP-style vision--language models and standard softmax classifiers under normalization, and extends to patch-token VLMs under mild architectural assumptions. While Lipschitzness ensures stability, it is the low effective dimensionality of the shared embedding space that governs the resulting sample complexity.

Our uniform convergence proof is based on covering-number or $(\varepsilon,\delta)$ arguments that rely on Lipschitz control of the evaluation functions with respect to the model parameters or representations. For CLIP-style VLMs and standard softmax classifiers, this Lipschitz property holds under mild normalization assumptions. CLIP (Contrastive Language--Image Pretraining; Radford et al., 2021) is a prominent VLM that learns a unified alignment space for language and visual features.

\section{Sample Complexity for Vision--Language Models}

\subsection{Uniform Convergence via Lipschitz Stability and $\varepsilon$-Covers (Model-Agnostic)}

Uniform convergence in finite-sample learning theory typically arises from a simple and general mechanism: stability under small parameter perturbations combined with finite coverings of the parameter space. When a family of evaluation functions varies smoothly with respect to a low-dimensional parameterization, uniform convergence can be obtained by discretizing the parameter space into finitely many $\varepsilon$-balls and applying concentration inequalities on this finite set.

\begin{figure}[t]
  \centering
  \includegraphics[width=0.9\linewidth]{LipschitzFigure1.png}
  \caption{\textbf{Geometrical intuition for uniform convergence.}
The horizontal axis denotes the prompt embedding $p$, and the vertical axis denotes the classifier score $f(p)$. The empirical function $f_n(p)$ approximates the true function $f(p)$ to within $\varepsilon$ uniformly across $p$. Because $f$ is Lipschitz with constant $L$, changes in the score are bounded by distance in embedding space, forming a Lipschitz tube around $f$. Covering the prompt space with balls of radius $\rho = O(\varepsilon/L)$ ensures that controlling error at a finite set of representative prompts suffices to control the error everywhere.}

  \label{fig:vlm_sample_complexity}
\end{figure}

\begin{figure}[t]
  \centering
  \includegraphics[width=0.5\linewidth]{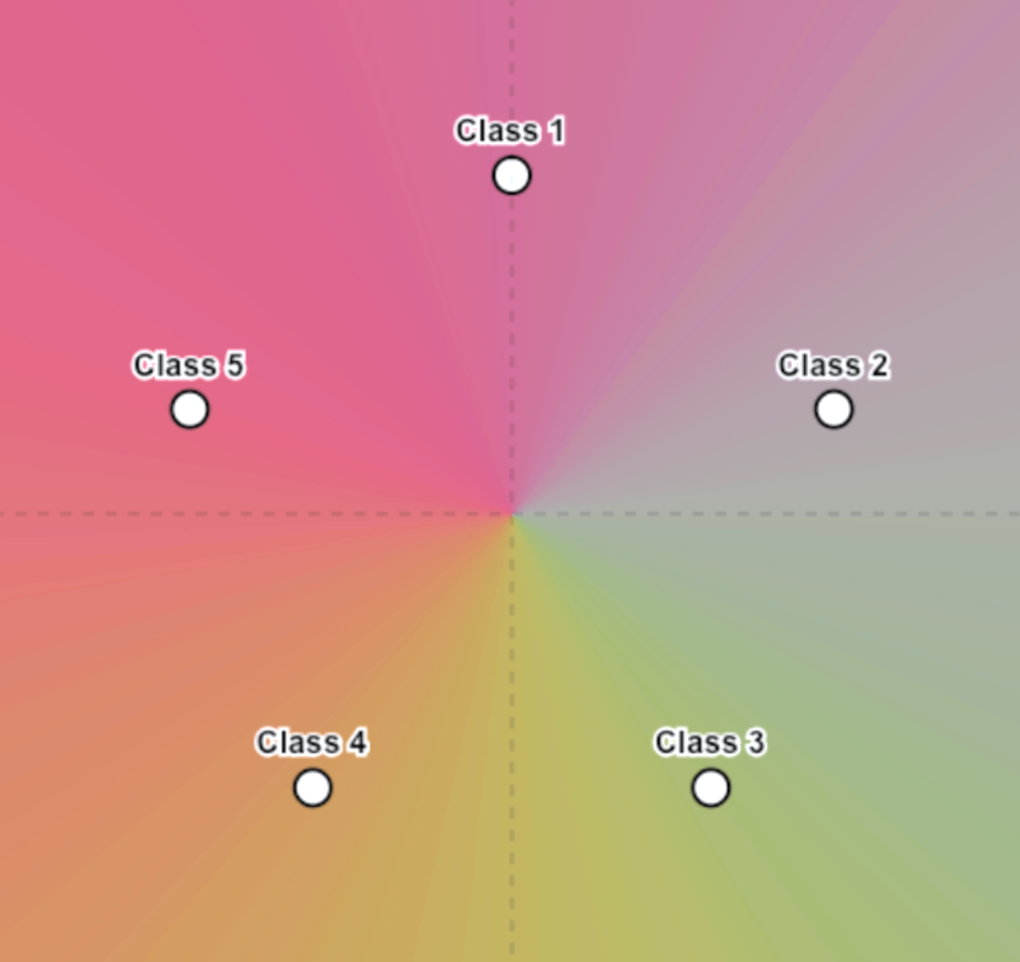}
  \caption{\textbf{Class probability outputs of a vision--language model (VLM) as a function of the prompt embedding.} The figure illustrates how predicted class probabilities vary smoothly across the semantic embedding space (shown in a 2D projection for visualization), enabling uniform control of accuracy and calibration under Lipschitz assumptions.}
  \label{fig:vlm_class_probs}
\end{figure}

Formally, let $\Theta$ be a parameter space equipped with a metric $d(\cdot,\cdot)$, and let
\[
\mathcal{F} = \{ f_\theta : \theta \in \Theta \}
\]
be a family of evaluation functionals (e.g., loss or calibration surrogates). If $f_\theta$ is Lipschitz in $\theta$, then uniform convergence over $\mathcal{F}$ follows by covering $\Theta$ with finitely many $\varepsilon$-balls.

\paragraph{Lemma 1 (Uniform convergence for Lipschitz families).}
Assume that for all $(x,y)$ and all $\theta,\theta' \in \Theta$,
\[
|f_\theta(x,y) - f_{\theta'}(x,y)| \;\le\; L\, d(\theta,\theta').
\]
Then for any $\varepsilon,\delta \in (0,1)$, there exists a universal constant $c>0$ such that, with probability at least $1-\delta$,
\[
\sup_{\theta \in \Theta}
\left|
\frac{1}{n}\sum_{i=1}^n f_\theta(X_i,Y_i)
-
\mathbb{E}[f_\theta(X,Y)]
\right|
\;\le\;
\varepsilon,
\]
provided that
\[
n \;\ge\; c\,\frac{1}{\varepsilon^2}
\left(
\log N(\Theta,\rho) + \log \frac{1}{\delta}
\right),
\]
where $N(\Theta,\rho)$ denotes the covering number of $\Theta$ at radius $\rho$.

Choosing $\rho = \varepsilon/L$ yields a sample complexity
\[
n(\varepsilon,\delta)
\;\ge\;
c\,\frac{1}{\varepsilon^2}
\left(
\log N(\Theta,\varepsilon/L) + \log \frac{1}{\delta}
\right).
\]

Thus, sample complexity is governed by the \emph{metric entropy} of the parameter space, not its ambient dimension. Metric entropy is the logarithm of the minimum number of $\varepsilon$-sized neighborhoods required to cover the model’s effective parameter space; it is a geometric notion and should not be confused with information-theoretic entropy.

\subsection{Specialization to Vision--Language Models: Lipschitzness in the Prompt Embedding}

We now specialize this generic mechanism to vision--language models. Let $z(x) \in \mathbb{S}^{D-1}$ denote a normalized image embedding and let $p_j \in \mathbb{S}^{D-1}$ denote the text embedding associated with disease class $j$. In CLIP-style models, class logits take the form
\[
s_j(x) = \alpha \langle z(x), p_j \rangle,
\]
and class probabilities are given by the softmax
\[
\pi_j(x) = \frac{\exp(s_j(x))}{\sum_k \exp(s_k(x))}.
\]

Under normalization, the logits are linear and hence globally Lipschitz in the prompt embedding with constant $\alpha$. The softmax map is smooth with bounded Jacobian, so each probability $\pi_j(x)$ is Lipschitz in the logits, and therefore Lipschitz in the prompt embeddings.

Moreover, commonly used evaluation functionals—such as cross-entropy loss, Brier score, or smoothed calibration indicators—are Lipschitz compositions of $\pi(x)$. Consequently, the induced evaluation class satisfies the Lipschitz condition of Lemma~1 with an explicit constant $L$ depending on the temperature $\alpha$ and, for calibration, a smoothing parameter.

In short, CLIP-style VLM classifiers are naturally Lipschitz with respect to prompt embeddings, requiring no additional architectural assumptions beyond normalization.

\subsection{Effective Dimension and Covering Numbers of the Prompt Space}

Lipschitzness alone does not determine sample complexity; the key quantity is the size of the $\varepsilon$-cover, which depends on the effective dimension of the parameter space. We assume that disease prompts lie in a low-dimensional semantic subspace,
\[
p_j \in \mathcal{M} \subset \mathbb{R}^D,
\qquad \dim(\mathcal{M}) = d \ll D.
\]
This assumption is empirically supported by spectral analyses of text embeddings, particularly in medical domains where disease concepts are highly correlated.

Under this model, the parameter space becomes
\[
\Theta = (\mathbb{S}^{D-1})^N.
\]
There exists a universal constant $c_1>0$ such that the covering number of $\Theta$ satisfies
\[
\log N(\Theta,\rho)
\;\le\;
c_1\, N d \log\!\left(\frac{1}{\rho}\right),
\]
independent of the ambient embedding dimension $D$.

Substituting this bound into Lemma~1, there exists a universal constant $c_2>0$ such that the sample complexity satisfies
\[
n(\varepsilon,\delta)
\;\ge\;
c_2\left[
\frac{N d}{\varepsilon^2}
\log\!\left(\frac{L}{\varepsilon}\right)
+
\frac{1}{\varepsilon^2}
\log\!\left(\frac{1}{\delta}\right)
\right].
\]

Thus, uniform convergence depends on the intrinsic semantic dimension $d$, rather than the ambient embedding dimension $D$. Vision--language models reduce effective complexity by inducing a shared, low-rank hypothesis class across diseases.

\subsection{Extensions and Remarks}

\subsubsection{Calibration and Binning}

Hard calibration bins are not Lipschitz due to discontinuities at bin edges. This issue is standard and can be addressed by either (i) replacing hard bins with smooth gating functions, or (ii) analyzing piecewise-Lipschitz behavior away from bin boundaries. Both approaches preserve the $\varepsilon$-cover argument and lead to the same dimension-dependent sample complexity.

\subsubsection{Patch-Token Vision--Language Models}

For transformer-based VLMs operating on sets of image and text tokens, logits are smooth compositions of dot products and attention weights. Under bounded token norms and standard softmax attention, the same Lipschitz arguments apply with respect to a low-dimensional text-token subspace, allowing Lemma~1 to be reused with modified constants.

\subsubsection{Spectral (Eigenvalue-Dependent) Sample Complexity Bounds}

The $\varepsilon$-cover argument can be extended when additional information about the spectrum of the text embedding space is available. Instead of assuming prompts lie in a strictly $d$-dimensional subspace, we model them as lying in an ellipsoid aligned with the eigenvectors of their empirical covariance:
\[
\mathcal{E} = \left\{ p \in \mathbb{R}^D : \sum_{i=1}^D \frac{p_i^2}{\lambda_i} \le 1 \right\},
\]
where $\lambda_1 \ge \lambda_2 \ge \cdots$ are eigenvalues. Directions with small $\lambda_i$ correspond to semantically negligible variations.

There exists a universal constant $c_3>0$ such that the metric entropy of the ellipsoid $\mathcal{E}$ at resolution $\rho$ satisfies
\[
\log N(\mathcal{E},\rho)
\;\le\;
c_3
\sum_i
\log\!\left(
\frac{\sqrt{\lambda_i}}{\rho}
\right)_+,
\]
where $(\cdot)_+ = \max\{\cdot,0\}$. Consequently, only eigen-directions satisfying $\sqrt{\lambda_i} \ge \rho$ contribute to the covering number.

Substituting this bound into the Lipschitz $\varepsilon$-cover argument, there exists a universal constant $c_4>0$ such that the sample complexity satisfies
\[
n(\varepsilon,\delta)
\;\ge\;
c_4\,\frac{1}{\varepsilon^2}
\left(
\sum_i
\log\!\left(
\frac{L\sqrt{\lambda_i}}{\varepsilon}
\right)_+
+
\log\!\frac{1}{\delta}
\right),
\]
where $L$ denotes the Lipschitz constant of the evaluation functional. For $N$ class prototypes, the summation term scales linearly with $N$.

This formulation shows that effective sample complexity is governed not by ambient dimension, but by spectral decay of the embedding space. Rapid eigenvalue decay implies that only a small number of directions are distinguishable at scale $\varepsilon/L$, yielding substantially lower data requirements.

\paragraph{Remark (DKW-style intuition in low intrinsic dimension; This argument is heuristic and intended to convey scaling behavior rather than a sharp distribution-free bound.).}
Although classical Dvoretzky--Kiefer--Wolfowitz bounds apply to one-dimensional distributions, they provide useful intuition for sample complexity in structured, low-dimensional settings.
When model outputs (e.g., classifier logits, posterior probabilities, or generative-model scores) depend in a Lipschitz manner on a low-dimensional latent variable of intrinsic dimension $r$, uniform deviations between empirical and population quantities scale as $O(\sqrt{r/n})$ up to logarithmic factors.
Consequently, achieving uniform accuracy or calibration tolerance $\varepsilon$ requires on the order of
\[
n \;\sim\; \frac{L^2 r}{\varepsilon^2}
\]
samples, where $L$ denotes the Lipschitz sensitivity of the model outputs to the latent representation.
This scaling explains why vision--language models can achieve reliable classification and calibration with moderate sample sizes despite operating in high-dimensional embedding spaces.

\subsubsection{Broader Interpretation}

Overall, our analysis clarifies why vision--language models can achieve strong accuracy and calibration with moderate amounts of data: uniform convergence depends on smoothness and intrinsic dimension of the induced representation, not directly on the number of model parameters, although model size may indirectly affect these quantities through its impact on stability and representation geometry. In this sense, vision--language models achieve favorable sample complexity not by avoiding classical uniform convergence theory, but by satisfying its conditions in a low-dimensional semantic space.

\section{Practical Implications}

The uniform convergence arguments developed here rely on Lipschitz control of logits, probabilities, and calibration surrogates with respect to the prompt embedding. This condition is naturally satisfied by CLIP-style vision--language models and standard softmax classifiers under normalization, and extends to patch-token VLMs under mild architectural assumptions. We do not claim that general VLM parameterizations satisfy Glivenko--Cantelli conditions. Rather, we analyze an induced low-dimensional family of classifiers obtained by varying prompts within a restricted semantic subspace, and derive finite-sample uniform convergence bounds for evaluation functionals on this restricted family.

The main practical consequence of this work is that finite-sample uniform convergence bounds provide guidance on when training models is statistically justified and when sufficient data have been collected. In biomedical settings, one often has access to on the order of $10^5$--$2\times 10^5$ medical images paired with text annotations, which can be further augmented using standard image and text augmentation techniques. However, such datasets remain orders of magnitude smaller than the hundreds of millions of image--text pairs used to train large foundation models such as CLIP. Our results show that, under appropriate structural assumptions, uniform convergence can still be achieved at these more modest sample sizes, thereby motivating the training of medical VLMs with currently available data.

The formal analysis shows that, under mild assumptions, one can either bound test error at a given sample size or determine the sample size required to bound error uniformly over the input domain. Although these bounds may appear lower than expected when compared to parameter-count heuristics, they provide theoretical support for training $N$-way diagnostic models of disease at current data scales, as well as for other conditional generative tasks such as style transfer and cross-modal image synthesis.

Conversely, researchers developing normative models for brain data—a specific class of generative models—should be aware that good fit according to standard calibration metrics, such as Expected Calibration Error (ECE) or Kullback--Leibler divergence, does not guarantee good worst-case behavior. This distinction is critical in clinical settings, where failures often occur in rare or extreme regimes. While ECE measures an average discrepancy across predefined confidence bins, classical results such as the Dvoretzky--Kiefer--Wolfowitz inequality control the maximum deviation between empirical and true distributions uniformly over all probability thresholds. As a result, a model may exhibit low ECE while remaining severely miscalibrated in rare or high-confidence regions. In this sense, ECE can be viewed as a coarse, binned approximation to the same underlying distributional mismatch that DKW controls uniformly.

Our analysis is motivated by the goal of understanding when DKW-style uniform guarantees might hold for model-induced families of predictions, rather than relying solely on average calibration metrics. We therefore encourage further work on uniform convergence bounds and on assessing calibration across the entire input domain, as also emphasized in recent discussions of model uncertainty and evaluation.

\subsection{Open Questions}

In future work, we will examine how non-uniform sampling affects the convergence bounds derived here. Approaches based on Stein thinning or Stein operators may offer promising avenues for addressing this problem. In addition, evaluation under non-i.i.d.\ test conditions or in the presence of covariate shift would generally weaken uniform guarantees of the type considered here. However, domain-shift mitigation techniques commonly used in generative modeling—such as flow matching or SHASH-based distributional models—may provide practical mechanisms for restoring calibration and stability in these settings.

\section*{Acknowledgments}

PMT is supported by NIH grants U01AG068057, RF1AG081571, R01MH134962, and R01MH131806.


\begin{thebibliography}{99}

\bibitem{Belhumeur1997}
Belhumeur, P.~N., Hespanha, J.~P., \& Kriegman, D.~J. (1997).
Eigenfaces vs.\ Fisherfaces: Recognition using class specific linear projection.
\emph{IEEE Transactions on Pattern Analysis and Machine Intelligence}, 19(7), 711--720.

\bibitem{Boecking2022}
Boecking, B., Usuyama, N., Bannur, S., Castro, D.~C., Schwaighofer, A., Hyland, S., et al. (2022).
Making the most of text semantics to improve biomedical vision--language processing.
In \emph{European Conference on Computer Vision} (pp.\ 1--21). Springer.

\bibitem{Dhinagar2025a}
Dhinagar, N.~J., Thomopoulos, S.~I., \& Thompson, P.~M. (2025).
Leveraging a vision--language model with natural text supervision for MRI retrieval, captioning, classification, and visual question answering.
\emph{bioRxiv}. https://doi.org/10.1101/2025.02.15.638446

\bibitem{Dhinagar2025b}
Dhinagar, N.~J., Ozarkar, S.~S., Buwa, K.~U., Thomopoulos, S.~I., Owens-Walton, C., Laltoo, E., et al. (2025).
Parameter-efficient fine-tuning of transformer-based masked autoencoders enhances resource-constrained neuroimage analysis.
\emph{bioRxiv}. https://doi.org/10.1101/2025.02.15.638442

\bibitem{DKW1956}
Dvoretzky, A., Kiefer, J., \& Wolfowitz, J. (1956).
Asymptotic minimax character of the sample distribution function and of the classical multinomial estimator.
\emph{Annals of Mathematical Statistics}, 27(3), 642--669.

\bibitem{Epstein2019}
Epstein, D., \& Meir, R. (2019).
Generalization bounds for unsupervised and semi-supervised learning with autoencoders.
\emph{arXiv preprint arXiv:1902.01449}.

\bibitem{Grisel2025}
Grisel, O. (2025).
Prediction intervals [tutorial].
\emph{PyData Paris 2025}.
Available at \url{https://www.youtube.com/watch?v=j0tcIwv6RBE}

\bibitem{Guo2017}
Guo, C., Pleiss, G., Sun, Y., \& Weinberger, K.~Q. (2017).
On calibration of modern neural networks.
In \emph{Proceedings of the 34th International Conference on Machine Learning} (pp.\ 1321--1330). PMLR.

\bibitem{KMT1975}
Koml\'os, J., Major, P., \& Tusn\'ady, G. (1975).
An approximation of partial sums of independent RVs and the sample DF. I.
\emph{Zeitschrift f\"ur Wahrscheinlichkeitstheorie und Verwandte Gebiete}, 32(1--2), 111--131.

\bibitem{KMT1976}
Koml\'os, J., Major, P., \& Tusn\'ady, G. (1976).
An approximation of partial sums of independent RVs and the sample DF. II.
\emph{Zeitschrift f\"ur Wahrscheinlichkeitstheorie und Verwandte Gebiete}, 34(1), 33--58.

\bibitem{Kumar2019}
Kumar, A., Liang, P.~S., \& Ma, T. (2019).
Verified uncertainty calibration.
In \emph{Advances in Neural Information Processing Systems 32} (pp.\ 3787--3798).

\bibitem{Massart1990}
Massart, P. (1990).
The tight constant in the Dvoretzky--Kiefer--Wolfowitz inequality.
\emph{Annals of Probability}, 18(3), 1269--1283.

\bibitem{Nayshabur2018}
Nayshabur, B., Sedghi, H., \& Zhang, C. (2018).
What is being transferred in transfer learning?
In \emph{Advances in Neural Information Processing Systems 31} (pp.\ 512--523).

\bibitem{Oymak2019}
Oymak, S., Fabian, Z., Li, M., \& Soltanolkotabi, M. (2019).
Generalization guarantees for neural networks via harnessing the low-rank structure of the Jacobian.
\emph{arXiv preprint arXiv:1906.05392}.

\bibitem{Radford2021}
Radford, A., Kim, J.~W., Hallacy, C., Ramesh, A., Goh, G., Agarwal, S., et al. (2021).
Learning transferable visual models from natural language supervision.
In \emph{Proceedings of the 38th International Conference on Machine Learning} (pp.\ 8748--8763). PMLR.

\bibitem{Turk1991}
Turk, M., \& Pentland, A. (1991).
Eigenfaces for recognition.
\emph{Journal of Cognitive Neuroscience}, 3(1), 71--86.

\bibitem{Vaicenavicius2019}
Vaicenavicius, J., Widmann, D., Andersson, C., Lindsten, F., Roll, J., \& Sch\"on, T.~B. (2019).
Evaluating model calibration in classification.
In \emph{Proceedings of the 22nd International Conference on Artificial Intelligence and Statistics} (pp.\ 3459--3467). PMLR.

\bibitem{Varoquaux2025}
Varoquaux, G. (2025).
Judging uncertainty from black-box classifiers [keynote].
Available at \url{https://www.youtube.com/watch?v=SI6bde9CKkc}

\bibitem{Wang2022}
Wang, Z., Wu, Z., Agarwal, D., \& Sun, J. (2022).
MedCLIP: Contrastive learning from unpaired medical images and text.
In \emph{Proceedings of the 2022 Conference on Empirical Methods in Natural Language Processing} (pp.\ 3876--3887). Association for Computational Linguistics.

\end{thebibliography}
\end{document}